%% file: main.tex
\documentclass{article}

\usepackage{arxiv}
\renewcommand{\headeright}{}
\renewcommand{\undertitle}{}

\usepackage[utf8]{inputenc}
\usepackage[T1]{fontenc}
\usepackage{microtype}
\usepackage{enumitem}

\usepackage{amsmath,amssymb,amsthm}
\usepackage{bm}
\usepackage{mathtools}

\usepackage{graphicx}
\usepackage{booktabs}
\usepackage{multirow}
\usepackage{makecell}
\usepackage{subcaption}

\usepackage{algorithm}
\usepackage{algpseudocode}

\usepackage[numbers,sort&compress]{natbib}

\usepackage{xcolor}
\usepackage{hyperref}
\usepackage{url}
\usepackage[capitalize,noabbrev]{cleveref}
\hypersetup{
  colorlinks=true,
  linkcolor=blue!60!black,
  citecolor=blue!60!black,
  urlcolor=blue!60!black,
}

\theoremstyle{definition}

\theoremstyle{remark}

\newcommand{\R}{\mathbb{R}}
\newcommand{\E}{\mathbb{E}}
\newcommand{\N}{\mathcal{N}}

\newcommand{\dif}{\mathrm{d}}
\newcommand{\dq}{\delta_{q}}
\newcommand{\dpath}{\delta_{\mathrm{path}}}

\title{Exact Global MCMC with Denoising Diffusion}

\author{%
  Mitch Hill\\
  \texttt{mitchhill128@gmail.com} \\
}

\date{} 

\begin{document}
\maketitle

\begin{abstract}
This work shows that diffusion models learned with standard denoising loss can provide effective global MCMC proposals for complex high-dimensional target densities. The method is motivated by the observation that sequentially applying a forward and reverse diffusion process defines a Markov chain with a target stationary distribution for an ideal denoiser trained on samples of the target distribution. This observation can be made exact for any denoiser by applying a Metropolis-Hastings step whose acceptance ratio includes the density of the forward and reverse paths of a discrete time SDE approximation. We therefore propose to train denoising diffusion models on locally convergent MALA samples to learn global MCMC proposals. We call the composition of the global denoiser-based path sampler and a local MALA sampler Denoising Diffusion Monte Carlo (DDMC). Experiments show that DDMC can provide global proposals with high acceptance across a variety of complex target densities. Our results offer preliminary evidence that the established scaling behavior of standard diffusion training transfers directly to exact sampling from high-dimensional unnormalized densities.
\end{abstract}

\input{sections/intro}

\input{sections/background}

\input{sections/method}

\input{sections/related}

\input{sections/experiments}

\input{sections/conclusion}

\section*{LLM Tool Disclosure}
Large language model assistants (ChatGPT 5.5 from OpenAI, and Claude Opus 4.8, Claude Opus 5, and Claude Fable 5 from Anthropic) were used during this project for writing and editing assistance, literature search, coding support, and checking of derivations and experimental bookkeeping. All research ideas, the method design, and the experimental decisions originated with the author, who reviewed and verified all content of the paper.

\bibliographystyle{unsrtnat}
\bibliography{references}

\appendix

\input{sections/appendix}

\end{document}

%% file: sections/intro.tex
\section{Introduction}
\label{sec:intro}

An ideal diffusion model is already an MCMC sampler. If $X$ is a draw from a target density $q$, then noising it to $Y = X + \sigma Z$ and denoising $Y$ with an ideal reverse process returns a fresh draw from $q$: the noise-to-denoise map leaves the target invariant. At large $\sigma$ this process can move freely between regions of the state space that no local sampler can connect. Sampling from unnormalized densities is a core computational problem in Bayesian inference and the physical sciences. Developing samplers that efficiently cover a multi-modal state space while closely adhering to the target density is a central challenge in this setting.

Since a trained diffusion model cannot be ideal, a practical sampler needs a way to account for modeling and discretization error in the reverse path. This can be accomplished by applying a natural Metropolis--Hastings (MH) step on the joint density of the discrete diffusion path, which makes the chain exact for any model of the reverse path. An inaccurate model costs acceptance, but the chain remains exact. 

We investigate training the reverse path model using a standard denoising diffusion loss, which in the ideal case learns a score model which fulfills exact transport from noised to the target density and has been found to be effective in many applications. Obtaining data for denoiser training raises a difficulty. Exact samples from the target are often unavailable, since producing them is the problem being solved. However, locally converged samples are easy to obtain: Metropolis-adjusted Langevin algorithm (MALA) chains launched from many initializations equilibrate within basins at low cost, even though the resulting corpus could weight those basins incorrectly. The same MH step that absorbs model error also absorbs this corpus bias. The path kernel and MALA are then natural partners: the diffusion path proposes global moves between basins that MALA cannot make, while the interleaved MALA steps continually return the chain to the distribution the denoiser was trained on, and in the limit of zero path acceptance the sampler degrades gracefully to plain MALA rather than failing. We call the composed sampler \emph{Denoising Diffusion Monte Carlo} (DDMC).

This training choice is where our approach departs from prior learned samplers, which predominantly train from the unnormalized density itself with variational path-space objectives \citep{zhang2022path, vargas2023denoising, vargas2024transport, guo2026proximal}. Metropolis--Hastings corrections of learned diffusion paths, trained variationally, were proposed concurrently and independently by MAD-Path \citep{chen2026madpath} and SPS \citep{chen2026stochastic}; we discuss the relationship in Section~\ref{sec:method}. Variational objectives carry two structural limits: the training signal comes from the model's own trajectories, so regions the model does not visit provide no gradient, and the objective differentiates through the sampling trajectory, so training memory grows with the number of noise levels. Denoising regression has neither limit and simply requires pairs of clean and noisy data. Sample-based learning is effective here because data is not the binding constraint: the unnormalized density is a data generator, and cheap, parallel local MCMC manufactures unlimited training samples. Denoising training is also the recipe with the best-established scaling behavior in machine learning \citep{dhariwal2021diffusion, peebles2023scalable, esser2024scaling}.

Our experiments support this design across Gaussian mixtures, particle potentials, and a $550$-dimensional Bayesian neural network posterior. DDMC matches the energy distribution of ground-truth samples at the measured estimator floor on every target, recovers correct mode weights from training corpora whose weights are significantly biased, and attains the best energy distances among diffusion-based samplers on the standard particle benchmarks. On the $550$-dimensional posterior, a $305$M-parameter denoiser yields global proposals accepted at rates of $0.13$ to $0.28$, and a cold-started chain reaches the equilibrium energy region within one or two proposal rounds. Our contributions are as follows:
\begin{itemize}[leftmargin=2em]
\item \textbf{DDMC}, an exact MCMC sampler whose global proposals come from a standard denoising diffusion model, composed with local MALA steps, requiring no change to the diffusion modeling or learning formulation.
\item \textbf{A practical end-to-end recipe}: corpus generation from cold starts via calibrated gradient ascent and MALA, moment calibration of the reverse kernel variances, a single-proposal acceptance diagnostic for monitoring training, and an i-SIR extension that amplifies acceptance and sampling efficiency by exploiting modern GPU throughput.
\item \textbf{Empirical validation} on targets up to $d = 550$, including exactness at measured ground-truth floors, repair of mis-weighted training corpora, and accurate modeling of the LJ-55 equilibrium, a benchmark on which prior diffusion-based samplers degrade sharply or fail to train.
\end{itemize}

%% file: sections/background.tex
\section{Background}
\label{sec:background}

\subsection{Sampling from Unnormalized Densities}
\label{sec:unnorm}

We consider the problem of drawing samples from a target density $q$ on $\R^d$ that is known only up to a normalizing constant. We assume we can evaluate a function $\tilde{q}(x)$ which satisfies
\begin{equation}
q(x) = \frac{\tilde{q}(x)}{\mathcal{Z}}, \qquad \mathcal{Z} = \int_{\R^d} \tilde{q}(x)\, \dif x,
\end{equation}
where the constant $\mathcal{Z}$ is intractable. We further assume access to the gradient $\nabla \log \tilde{q} = \nabla \log q$, often called the score, which is unaffected by the unknown constant. This setting covers Bayesian posteriors, where $\mathcal{Z}$ is the marginal likelihood, and Boltzmann distributions $\tilde{q}(x) = e^{-U(x)}$ from statistical physics, where $\mathcal{Z}$ is the partition function.

\subsection{Metropolis--Hastings Correction and Local Samplers}
\label{sec:mh}

Markov chain Monte Carlo (MCMC) draws samples from $q$ by simulating a Markov chain whose stationary distribution is $q$. Given any proposal kernel $R(x' \mid x)$, the Metropolis--Hastings (MH) correction accepts a proposed move $x \rightarrow x'$ with probability
\begin{equation}
\alpha(x, x') = \min\left\{ 1,\; \frac{\tilde{q}(x')\, R(x \mid x')}{\tilde{q}(x)\, R(x' \mid x)} \right\},
\label{eq:mh}
\end{equation}
and otherwise leaves the chain at $x$ \citep{metropolis1953equation, hastings1970monte}. The ratio requires only the unnormalized density, and the corrected kernel satisfies detailed balance with respect to $q$, so $q$ is stationary for any choice of $R$. The proposal affects how quickly the chain mixes.

The Metropolis-adjusted Langevin algorithm (MALA) \citep{roberts1996exponential} uses the proposal
\begin{equation}
x' = x + \frac{h^2}{2} \nabla \log q(x) + h\, \xi, \qquad \xi \sim \N(0, I_d),
\label{eq:mala}
\end{equation}
that is, $R_h(x' \mid x) = \N\!\left(x';\, x + \tfrac{h^2}{2} \nabla \log q(x),\, h^2 I_d\right)$, a single Euler--Maruyama step of the overdamped Langevin diffusion, followed by the MH correction \eqref{eq:mh}. In this parameterization $h$ is the standard deviation of the proposal noise, and for well-tuned MALA it sits near the scale of the smallest local standard deviation of the target, so each move is small relative to the global structure of the target. As a result MALA mixes well within a single mode but must cross low-density barriers through many small steps, and the expected time to move between well-separated modes grows exponentially with the barrier height. Gradient-based samplers with longer coherent moves, such as Hamiltonian Monte Carlo \citep{duane1987hybrid, neal2011mcmc} and NUTS \citep{hoffman2014no}, improve local exploration but share the same limitation: their trajectories follow the energy landscape and do not cross regions of negligible density, so multimodal targets remain out of reach.

\subsection{Denoising Diffusion Models}
\label{sec:diffusion}

We summarize denoising diffusion in the EDM formulation of \citet{karras2022elucidating}. Diffusion models operate on the family of smoothed densities $q_\sigma = q * \N(0, \sigma^2 I_d)$ across noise levels $\sigma \in [\sigma_{\min}, \sigma_{\max}]$, where $q_{\sigma_{\min}} \approx q$ and $q_{\sigma_{\max}}$ is close to the tractable Gaussian $\N(0, \sigma_{\max}^2 I_d)$ when $\sigma_{\max}$ is large relative to the scale of the data. A denoiser network $D_\theta(y, \sigma)$ is trained by denoising regression on samples $X \sim q$:
\begin{equation}
\mathcal{L}(\theta) = \E_{\sigma}\left[ \lambda(\sigma)\, \E_{X \sim q,\, Z \sim \N(0, I_d)} \left\| D_\theta(X + \sigma Z, \sigma) - X \right\|^2 \right],
\label{eq:dsm}
\end{equation}
with a weighting $\lambda(\sigma)$ and a training distribution over noise levels. The minimizer of \eqref{eq:dsm} is the posterior mean $D^*(y, \sigma) = \E[X \mid X + \sigma Z = y]$, which is related to the score of the smoothed density by Tweedie's identity \citep{efron2011tweedie},
\begin{equation}
\nabla_y \log q_\sigma(y) = \frac{D^*(y, \sigma) - y}{\sigma^2},
\end{equation}
so a trained denoiser estimates the score of $q_\sigma$ at every noise level simultaneously. Training is a simulation-free regression where each update noises a data sample at a single level.

\paragraph{SDE Sampling and forward/backward path densities.}
Samples can be generated with a discrete-time SDE approximation in the style of DDPM ancestral sampling \citep{ho2020denoising, song2021score}, whose path densities are available in closed form. Let $x_0, x_1, \dots, x_T$ be states at the ladder levels $\sigma_{\min} = \sigma_0 < \sigma_1 < \cdots < \sigma_T = \sigma_{\max}$ and let $\Delta_k^2 = \sigma_k^2 - \sigma_{k-1}^2$ denote the variance added at level $k$. Continuous-time denoising training enables the use of any noise schedule $\{\sigma_k\}_{k=0}^{T}$ and any number of steps $T$. The forward process perturbs a clean sample $x_0$ with independent Gaussian increments,
\begin{equation}
p(x_k \mid x_{k-1}) = \N\!\left(x_k;\, x_{k-1},\, \Delta_k^2 I_d\right),
\label{eq:fwdkernel}
\end{equation}
so that a draw $x_0 \sim q$ followed by the forward increments \eqref{eq:fwdkernel} has the joint path density
\begin{equation}
p(x_{0:T}) = q(x_0)\, p(x_{1:T} \mid x_0) = q(x_0) \prod_{k=1}^{T} p(x_k \mid x_{k-1}).
\label{eq:fwdpath}
\end{equation}
The increments telescope, so $x_k \mid x_0 \sim \N\!\left(x_0,\, (\sigma_k^2 - \sigma_0^2)\, I_d\right)$: for $x_0 \sim q$ the forward process reproduces the smoothed marginals, $x_k \sim q_{\sigma_k}$ up to the negligible $\sigma_0 = \sigma_{\min}$.

Generation starts from $x_T \sim \N(0, \sigma_{\max}^2 I_d)$ and runs the ladder in reverse using Gaussian transition kernels built from the denoiser, with one evaluation of $D_\theta$ per level,
\begin{equation}
p_\theta(x_{k-1} \mid x_k) = \N\!\left(x_{k-1};\, \mu_\theta(x_k, \sigma_k),\, \tau_k^2 I_d\right),
\qquad
\mu_\theta(x_k, \sigma_k) = \alpha_k x_k + (1 - \alpha_k)\, D_\theta(x_k, \sigma_k),
\label{eq:revkernel}
\end{equation}
with $\alpha_k = \sigma_{k-1}^2 / \sigma_k^2$. The mean $\mu_\theta$ is the exact posterior mean of $x_{k-1}$ given $x_k$ when the unknown clean sample is replaced by the denoiser estimate $D_\theta(x_k, \sigma_k)$. The reverse variance $\tau_k^2$ is a modeling choice: the DDPM posterior variance $\tau_k^2 = \sigma_{k-1}^2 \Delta_k^2 / \sigma_k^2$ and the forward-matched variance $\tau_k^2 = \Delta_k^2$ are standard, and any positive choice yields a valid sampler. The backward path density is
\begin{equation}
p_\theta(x_{0:T}) = \N\!\left(x_T;\, 0,\, \sigma_{\max}^2 I_d\right) \prod_{k=1}^{T} p_\theta(x_{k-1} \mid x_k).
\label{eq:revpath}
\end{equation}
We highlight the property that makes these objects useful beyond generation: both the forward path density \eqref{eq:fwdpath} and the backward path density \eqref{eq:revpath} are products of Gaussian kernels whose means and variances are available in closed form, so the joint density of any given trajectory can be evaluated exactly in either direction with $T$ evaluations of $D_\theta$. No marginal density $q_{\sigma_k}$ ever needs to be computed.

%% file: sections/method.tex
\section{Method}
\label{sec:method}

The approach in this work is motivated by an observation about the properties of an ideal diffusion model. Consider a target density $q$ over $\mathbb{R}^d$ and let $q_\sigma$ be the convolution of $q$ with the Gaussian $\N(0, \sigma^2 I_d)$. A diffusion model defines a conditional distribution $\rho(x | y, \sigma)$ over endpoints $x$ of the reverse diffusion path starting from $y$ at noise level $\sigma$. The diffusion model is ideal if it transports samples from $q_\sigma$ exactly to $q$: if $Y\sim q_\sigma$ and $\hat{X} \sim \rho(\cdot | Y, \sigma)$, then $\hat{X} \sim q$. The distribution $\rho(\cdot | y, \sigma)$ is a density on $\mathbb{R}^d$ when the reverse path is an SDE and is a Dirac delta when it is a deterministic ODE. Let $X \sim q$. Sampling a forward path endpoint $Y = X + \sigma Z$ where $Z \sim \N(0, I_d)$ yields $Y \sim q_\sigma$. Sampling the reverse path endpoint $\hat{X} \sim \rho(\cdot | Y, \sigma)$ defines a Markov transition $X \rightarrow \hat{X}$. If $\rho$ is the conditional distribution of an ideal diffusion model, then $\hat{X} \sim q$. Therefore, for any $\sigma > 0$, $X \rightarrow \hat{X}$ defines an ergodic Markov chain with stationary distribution $q$, where ergodicity follows from the fact that the forward pass has full support and the reverse paths preserve support in both the SDE case and ODE case under mild regularity conditions (see Appendix~\ref{app:ergodicity} for more details). Large values of $\sigma$ should lead to global transitions which are capable of crossing the energy barriers which prevent local samplers from mixing between modes.

This natural connection between diffusion modeling and Markov chain sampling prompts the question which is the focus of this work: can denoising diffusion models be used as MCMC samplers which enable efficient global proposals? A variety of prior works have adopted ideas from diffusion modeling to propose diffusion-inspired samplers that differ from the standard denoising diffusion setting \citep{akhoundsadegh2024idem, guo2026proximal, zhang2022path, vargas2023denoising, berner2024optimal, vargas2024transport, huang2024reverse, phillips2024particle, chen2024diffusive, grenioux2024stochastic}. In this work, we show that the standard version of denoising diffusion typical of image modeling and related applications can provide an exact and powerful MCMC sampler without any changes to the modeling or learning formulation.

In practice the diffusion model cannot be ideal and an exact sampler must account for imperfections in the reverse path. This can be accomplished using a Metropolis--Hastings correction on the joint density of a discrete time diffusion path which approximates the SDE solution. We start with the situation discussed above. Let $X_0 \sim q$, and sample a joint forward path $(X_0, X_1, \dots, X_{T-1}, X_T)$ according to \eqref{eq:fwdpath}, which yields $X_T \sim q_{\sigma_T}$ as before. Then one can use \eqref{eq:revkernel} to sample a reverse path $(\hat{X}_0, \hat{X}_1, \dots, \hat{X}_{T-1}, X_T)$ from the shared top point $X_T$. A Metropolis--Hastings correction can be applied to the conditional distribution of $V = (X_0, X_1, \dots, X_{T-1})$ given $X_T$ as follows. By Bayes' rule, the conditional density of $V$ given $X_T$ is
\begin{equation}
p(v \mid x_T) = p(x_{0:(T-1)} \mid x_T) = \frac{p(x_{0:T})}{p(x_T)} \propto q(x_0) \prod_{k=1}^{T} p(x_k \mid x_{k-1})
\label{eq:fwdpathminusT}
\end{equation}
which follows from \eqref{eq:fwdpath}. The marginal density $p(x_T)$ is intractable, but $x_T$ is held fixed by the move, so it is a constant that will cancel in the MH ratio, as does the normalizing constant of $q$. The proposed path is drawn from the reverse kernels started at the shared top point $X_T$, so the Markov transition from $V$ to $\hat{V} = (\hat{X}_0, \hat{X}_1, \dots , \hat{X}_{T-1})$ has a kernel that does not depend on $V$:
\begin{equation}
R(\hat{v} \mid v) = p_\theta(\hat{x}_{0:(T-1)} \mid x_T) = \prod_{k=1}^T p_\theta (\hat{x}_{k-1} \mid \hat{x}_k), \qquad \hat{x}_T = x_T,
\label{eq:revpathminusT}
\end{equation}
which follows from \eqref{eq:revpath} with the initial Gaussian factor removed since $X_T$ is fixed. Plugging \eqref{eq:fwdpathminusT} and \eqref{eq:revpathminusT} into the Metropolis--Hastings acceptance probability \eqref{eq:mh} gives
\begin{equation}
\alpha(v, \hat{v}) = \min\left\{ 1,\; \frac{\tilde{q}(\hat{x}_0) \prod_{k=1}^{T} p(\hat{x}_k \mid \hat{x}_{k-1})  \prod_{k=1}^T p_\theta (x_{k-1} \mid x_k) }{\tilde{q}(x_0)\prod_{k=1}^{T} p(x_k \mid x_{k-1})  \prod_{k=1}^T p_\theta (\hat{x}_{k-1} \mid \hat{x}_k)} \right\}, \qquad \hat{x}_T = x_T,
\label{eq:mhpath}
\end{equation}
which ensures that $(\hat{X}_0, \hat{X}_1, \dots, \hat{X}_{T-1}, X_T)$ follows the forward path density \eqref{eq:fwdpath} since $(X_0, X_1, \dots, X_{T-1}, X_T)$ also does by construction: the MH step preserves the conditional distribution given $X_T$, and $X_T$ itself is untouched. The marginal of \eqref{eq:fwdpath} for $X_0$ is $q$, so that $\hat{X}_0 \sim q$. This construction provides a Markov transition $X_0 \rightarrow \hat{X}_0$ which has stationary distribution $q$. The chain is ergodic because the composite transition has strictly positive density: the forward increments and reverse kernels are full-support Gaussians, and the acceptance probability is positive wherever $\tilde{q} > 0$ (Appendix~\ref{app:ergodicity}). An interesting part of this construction is that the Markov chain is an exact sampler for $q$ regardless of the quality of the learned diffusion model $p_\theta$, with the caveat that a poorly learned $p_\theta$ will lead to a frozen chain.

We note that the Metropolis--Hastings step \eqref{eq:mhpath} was proposed in closely related forms by MAD-Path \citep{chen2026madpath} and SPS \citep{chen2026stochastic} concurrently with and independently of the development of this work. These works use a variational objective to train the score function using the target density itself rather than samples from the target density. This works well for targets with simpler geometry but fails to scale to more complex targets (see Section~\ref{sec:experiments}). The central contribution of this work is to show that sample-based training of a diffusion denoiser allows us to achieve global proposals with high acceptance for complex target distributions using \eqref{eq:mhpath}. By situating exact MCMC within the denoising diffusion paradigm, which is known to scale well with data, model size, and compute \citep{dhariwal2021diffusion, rombach2022high, saharia2022photorealistic, peebles2023scalable, esser2024scaling}, our work offers preliminary evidence that conventional techniques for scaling diffusion models can be naturally transferred to exact sampling from complex high-dimensional unnormalized densities.

\subsection{Denoising Diffusion Monte Carlo (DDMC)}

Since we choose to train $p_\theta$ using the denoising objective \eqref{eq:dsm}, the main design choice for learning the model is choosing how to obtain samples for training. In this work $p_\theta$ is trained only from samples without any direct knowledge of the target density. Our choice of dataset is motivated by two simple observations. The first is that an unnormalized density is a data generator. It is straightforward to use local MCMC samplers such as MALA to generate an unlimited number of samples which have locally converged in a certain region of the state space. Dataset size can be scaled simply by drawing more samples. The caveat of such a procedure is that regions of the state space which local samplers do not explore will be inaccessible for a diffusion proposal and that the weighting of a corpus of locally converged samples across modes might not represent the ground-truth weighting of modes across the target. The first caveat is strict and shared by all learned neural samplers. The second caveat can be alleviated by MH correction. Since \eqref{eq:mhpath} is exact for any $p_\theta$, this suggests that misweighted modes in the training set can be accommodated at the cost of acceptance provided the misweighting is not too extreme.

Based on these observations, we propose \emph{Denoising Diffusion Monte Carlo} (DDMC): an exact MCMC sampler whose global moves are the path-space MH step \eqref{eq:mhpath}, driven by a denoiser trained with the standard objective \eqref{eq:dsm} on a corpus of locally converged MALA samples, and whose local moves are MALA steps. Large corpora of locally converged samples can be obtained at a reasonable computational cost. The procedure we use to create a corpus is described in Section~\ref{sec:corpus}. The DDMC transition kernel is the composition of the diffusion MH step \eqref{eq:mhpath} and the MALA kernel:
\begin{equation}
K_{\mathrm{DDMC}}(x' \mid x)
\;=\;
\int_{\R^d} K_{\mathrm{path}}(u \mid x)\, K_h^{N}(x' \mid u)\, \dif u.
\label{eq:ddmc}
\end{equation}
$K_{\mathrm{path}}$ is the transition kernel of the bottom path endpoint under one path move: starting from $x_0 = x$, draw the forward path $x_{1:T} \sim p(\cdot \mid x_0)$, propose a reverse path from the shared top point, and accept or reject with \eqref{eq:mhpath}, so the new state is $\hat{x}_0$ on acceptance and $x_0$ otherwise. $K_h^{N}$ is the $N$-fold composition of the MALA kernel $K_h$ from Section~\ref{sec:mh}. Upon rejection each kernel places its mass at the current state. Since the path move and every MALA step leave $q$ invariant individually, the composition \eqref{eq:ddmc} leaves $q$ invariant.

The composition allows the strengths of each kernel component to support the weaknesses of the other. The denoiser is trained on MALA-generated samples, and the MALA steps of each cycle return the chain towards the distribution on which the denoiser was trained. If the path proposal is rejected many times in a row, the MALA steps continue to refine the local state, keeping the chain in configurations the denoiser handles capably. In the limit of zero path acceptance DDMC reduces to plain MALA. The path kernel enables global jumps across basins that are not possible with MALA. This allows the composite sampler to traverse modes where local MALA samples would be trapped indefinitely.

\subsection{Creating a MALA Corpus}
\label{sec:corpus}

This section describes a generic procedure for creating a large corpus of MALA samples to train a DDMC model. There are two main phases: 1) tuning the MALA step size and discovering the energy spectrum of locally converged MALA states and 2) tuning a producer algorithm to be able to generate new locally converged MALA samples as quickly as possible.

In our experiments, all MALA samples used for a training corpus are initialized from either a normal or uniform distribution that spans the high-probability regions of the target density. This choice is made because for the densities in our experiments, we found either that this initialization spanned all modes with a reasonable number of samples in each mode for learning the diffusion, or that the density had an overall unimodal structure where initialization did not affect the regions that MALA was able to explore. This choice has a clear failure case: if the target probability mass is concentrated in a region that cannot be accessed with this initialization, then learning will fail. Addressing this limitation is an important direction left for future work. One possible direction is to use techniques for mapping the basin structure of a target density \citep{doye1999double} and to initialize MALA from different parts of the global landscape structure.

The first phase calibrates MALA and measures the energy spectrum of locally converged states. From the cold initialization, we run gradient ascent on $\log \tilde{q}$ with Adam until convergence, which locates a set of modes. The MALA step size is then tuned at these modes: we select the largest $h$ whose worst-chain acceptance rate remains within a target band, since curvature differs across modes and a step size tuned on the average acceptance can be far too large for the sharpest basin. Finally, we run MALA from the modes with a per-chain adaptive step size until the mean of $\log \tilde{q}$ reaches a plateau. This plateau serves two purposes. Its quantiles define a reference energy band that all subsequent convergence checks are measured against, and the adapted step sizes at the plateau provide the sampling step size used for all later MALA legs.

The second phase converts this calibration into the cheapest fixed schedule that produces new locally converged samples from the cold initialization. The producer runs gradient ascent followed by MALA, and its one free parameter is how far the ascent is allowed to overshoot the reference band before MALA takes over: the ascent stops at the level $L(f) = L_{\mathrm{band}} + f\,(L_{\mathrm{mode}} - L_{\mathrm{band}})$ for $f \in [0, 1]$, where $f = 0$ stops the moment the band's energy is reached and $f = 1$ descends fully to the mode. Both extremes are defensible a priori: a small $f$ is cheap but a gradient trajectory point at the correct energy is not yet a thermal sample and may not have settled into a basin, while a large $f$ guarantees a genuine mode but pays for the thermal climb twice. We therefore sweep $f$, running each candidate schedule to the plateau and declaring it valid only if no chain freezes (acceptance above a floor for every chain) and the final energy quantiles match the reference band within tolerance. The cheapest valid schedule, measured in total gradient evaluations per sample, is selected, and its fixed step counts are re-validated on a fresh batch before being executed blindly at scale to produce the corpus.

\subsection{Diagnosing the Learning Procedure}
\label{sec:diagnostics}

It is essential to be able to quickly and accurately monitor whether a learned denoiser is a viable MCMC sampler. Having a simple and accurate metric drives decisions about data size, model architecture and size, and the amount of training steps required. In the case of an MH sampler like DDMC, the most important metric is acceptance probability for samples close to equilibrium. However, it is costly to roll out long MCMC trajectories to measure the acceptance probability. Since MALA corpus samples withheld from the training set can be interpreted as approximate equilibrium samples, a single-step proposal gives a reasonable estimate of the path kernel performance during long MCMC runs. The primary metric to diagnose learning is the expected acceptance probability of a single path proposal launched from a held-out state,
\begin{equation}
\bar{\alpha} \;=\; \E\big[ \min\{1, e^{\log r}\} \big],
\label{eq:diagacc}
\end{equation}
where $\log r$ is the logarithm of the MH ratio in \eqref{eq:mhpath} and the expectation runs over a held-out state $x_0$, its forward path, and one reverse proposal from the shared top point. The log-acceptance ratio decomposes into a target term and a path term,
\begin{align}
\log r(v, \hat{v}) &= \dq + \dpath, \label{eq:logr}\\
\dq &= \log \tilde{q}(\hat{x}_0) - \log \tilde{q}(x_0), \label{eq:dq}\\
\dpath &= \sum_{k=1}^{T} \Big[ \log p(\hat{x}_k \mid \hat{x}_{k-1}) - \log p(x_k \mid x_{k-1}) \Big]
+ \sum_{k=1}^{T} \Big[ \log p_\theta(x_{k-1} \mid x_k) - \log p_\theta(\hat{x}_{k-1} \mid \hat{x}_k) \Big], \label{eq:dpath}
\end{align}
with the convention $\hat{x}_T = x_T$ as before.

The term $\dq$ measures whether the endpoint proposal improves on the current state under the target, and $\dpath$ compares how well the forward and reverse processes from the starting point and proposal are explained by the learned model. A balance between $\dpath$ and $\dq$ is needed for chains to move across the entire energy spectrum of equilibrium samples rather than being driven only by $\dq$ and freezing at high $\log q$. The behavior of \eqref{eq:diagacc} is governed by the right tail of the distribution of $\log r$. To contextualize acceptance, it is helpful to look at both the mean and standard deviation of $\log r$ across input states and reverse path stochasticity and to visualize a histogram of its distribution. A reasonable acceptance ratio requires a significant amount of probability mass above 0 or within a small range of nats below 0. Both $\dq$ and $\dpath$ are sensitive to parameters of the reverse path, which is discussed in the next section. Histograms of $\dq$, $\dpath$, and $\log r$, together with the evolution of their means and of the acceptance \eqref{eq:diagacc} over training, are presented for each target in Appendix~\ref{app:diagnostics}.

\subsection{Tuning the Reverse Path}

There are four parameters which require proper tuning to achieve good acceptance rates using \eqref{eq:mhpath}. These are $\sigma_\textrm{max}$ and $\sigma_\textrm{min}$ which are the noise levels of the SDE at the top and bottom respectively of the reverse path, $T$ which is the number of noise levels in the reverse path, and $n_\textrm{cal}$ which gives the number of calibration samples used to tune the variances of the reverse path.

The parameter $\sigma_\textrm{max}$ affects the ability to propose global moves. If this value is too small, jumps between remote regions of the state space are not possible. An effective value of $\sigma_\textrm{max}$ must be large enough to bridge the state space distance between distinct modes. At noise levels far above the data scale the true reverse transitions are close to Gaussian regardless of model quality, so these levels contribute little to $\dpath$. Setting a value of $\sigma_\textrm{max}$ that is larger than necessary is not a major risk as long as $T$ is sufficiently large.

The parameter $\sigma_\textrm{min}$ affects the ability of the reverse path to propose samples which respect the local geometry of the target density. This value must be small enough to land within the thermal shell of the target density. A value of $\sigma_\textrm{min}$ that is too small pays a penalty on the $\dpath$ term because errors in reverse path values at low noise are amplified by a factor $1/\sigma^2$. Setting $\sigma_\textrm{min}$ to a value that is a few times smaller than the optimal MALA step size is a good initial choice. Sweeping over values in this range and observing the acceptance probability \eqref{eq:diagacc} can provide a more precise value. We find that the optimal $\sigma_\textrm{min}$ can be determined using a moderate $T$ and the same value is often effective for larger $T$.

The parameter $T$ affects the ability of the diffusion to accurately model the continuous time SDE of the reverse path. We did not observe risk using large $T$ beyond computational cost provided enough calibration samples are used (see below). The value of $T$ that is required depends on the complexity of the problem. Simpler targets work with smaller $T$ such as 80 or lower while more complex targets benefit from larger $T$ of 320 or above. It is straightforward to measure the effect of $T$ on $\log r$ and choose the smallest $T$ where the benefit to $\log r$ begins to saturate.

Calibration of the reverse kernel variances plays a critical role in achieving high acceptance. Recall from Section~\ref{sec:diffusion} that exactness holds for any choice of the reverse variances $\tau_k^2$, so calibration affects only the acceptance rate, never correctness. Given a calibration set of $n_\textrm{cal}$ states $\{x_0^{(i)}\}_{i=1}^{n_\textrm{cal}}$, we simulate one forward path from each state and set the variance at each level to the mean squared residual of the reverse mean,
\begin{equation}
\hat{\tau}_k^2 \;=\; \frac{1}{n_\textrm{cal}\, d} \sum_{i=1}^{n_\textrm{cal}} \left\| x_{k-1}^{(i)} - \mu_\theta\big(x_k^{(i)}, \sigma_k\big) \right\|^2,
\label{eq:taucal}
\end{equation}
which is the moment-matched variance and is the KL-optimal choice for a Gaussian kernel with mean $\mu_\theta$. The calibration states form a corpus that is separate from both the training set and the held-out states used for the diagnostic \eqref{eq:diagacc}: calibrating on training states would allow memorized residuals to shrink $\hat{\tau}_k$, while calibrating on the diagnostic states would bias the diagnostic itself.

The calibration set must be large enough, but the returns to additional calibration data vanish. A mis-calibrated $\tau_k$ can be costly: any departure from the optimal variance raises the per-level cost in $\dpath$, so estimation noise does not average out across levels but instead accumulates as a downward bias in $\log r$ that grows with the number of levels. Too few calibration states therefore depress acceptance in a way that might be mistaken for a deficiency of the learned model. Beyond a sufficient size, however, additional calibration data leaves $\log r$ essentially unchanged. In practice we determine the required $n_\textrm{cal}$ at a moderately large $T$ by increasing it until the acceptance \eqref{eq:diagacc} stops improving, and then use that value for larger $T$. Finally, calibration is a one-time cost: for a given denoiser and reverse path configuration $(T, \sigma_\textrm{max}, \sigma_\textrm{min})$, the calibrated variances $\hat{\tau}_{1:T}^2$ can be cached and reused for all subsequent sampling runs and evaluations.

\subsection{i-SIR Parallel Proposals to Amplify Acceptance}

The path move considered so far evaluates a single proposal per cycle, so at low acceptance rates most reverse passes are spent on rejected paths. A key structural property of the proposal kernel \eqref{eq:revpathminusT} enables a multiple-proposal extension: $R(\hat{v} \mid v)$ depends only on the shared top point $X_T$ and not on the current path, so any number of proposals can be drawn independently from the same distribution. This is the setting of iterated sampling importance resampling (i-SIR) \citep{andrieu2018uniform, samsonov2022local}, closely related to multiple-try Metropolis \citep{liu2000multiple}.

Given the current path $v$ and top point $x_T$, we set $\hat{v}^{(1)} = v$ and draw $K - 1$ independent reverse paths $\hat{v}^{(2)}, \dots, \hat{v}^{(K)} \sim R(\cdot \mid x_T)$ to form a pool of $K$ candidates. Each candidate receives an unnormalized importance weight equal to the ratio of the conditional target \eqref{eq:fwdpathminusT} to the proposal density,
\begin{equation}
w^{(j)} \;=\; \frac{\tilde{q}\big(\hat{x}_0^{(j)}\big) \prod_{k=1}^{T} p\big(\hat{x}_k^{(j)} \mid \hat{x}_{k-1}^{(j)}\big)}{\prod_{k=1}^{T} p_\theta\big(\hat{x}_{k-1}^{(j)} \mid \hat{x}_k^{(j)}\big)}, \qquad \hat{x}_T^{(j)} = x_T,
\label{eq:isirweight}
\end{equation}
and the next state is selected from the pool with probability proportional to $w^{(j)}$. Because the current state is included in the pool and the weights are the exact target-to-proposal ratios, this transition leaves the conditional target invariant by the same argument as the single-proposal case, and exactness again holds for any $p_\theta$. The probability of leaving the current state, $1 - w^{(1)} / \sum_{j} w^{(j)}$, is nondecreasing in $K$, so i-SIR amplifies the movement rate of the path kernel without changing its stationary distribution.

The practical motivation for $K > 2$ is hardware throughput. MCMC is inherently sequential, and a single reverse pass evaluates the denoiser on one state per noise level, which typically leaves a modern GPU far below saturation. The $K - 1$ proposals are mutually independent draws from the same distribution, so they can be generated in one batched reverse pass: until the batch dimension saturates the device, the wall-clock cost of a cycle with $K$ candidates is close to that of a single proposal, while the chance that the pool contains a high-weight candidate grows with $K$.

%% file: sections/related.tex
\section{Related Work}
\label{sec:related}

\paragraph{Exact path-space corrections.}
Accepting or rejecting an entire trajectory to obtain exact samples has a long ancestry. Tempered transitions \citep{neal1996sampling} apply a single acceptance test to a full sequence of annealed moves. The augmented forward/reverse path density at the heart of \eqref{eq:mhpath} appears in PDDS \citep{phillips2024particle}, where it defines sequential Monte Carlo importance weights rather than a Metropolis--Hastings test. Concurrently with this work, MAD-Path \citep{chen2026madpath} and SPS \citep{chen2026stochastic} propose Metropolis--Hastings corrections of learned reverse diffusion paths in a closely related form. Both train the score variationally from the target density. In this work, score models are trained from MALA samples with the standard denoising objective.

\paragraph{Variational diffusion samplers.}
A large family of methods trains a diffusion or controlled SDE to sample an unnormalized density by minimizing a path-space divergence, including PIS \citep{zhang2022path}, DDS \citep{vargas2023denoising}, the log-variance variant \citep{richter2024improved}, the optimal-control formulation of \citet{berner2024optimal}, CMCD \citep{vargas2024transport}, SCLD \citep{chen2025sequential}, adjoint-based samplers \citep{havens2025adjoint, liu2025adjoint}, and PDNS \citep{guo2026proximal}. These objectives share a structural property that bears on scaling: the training signal comes from trajectories simulated by the current model, so regions the model does not visit provide no gradient. Most members additionally differentiate through the sampling trajectory, so training memory grows linearly in the number of noise levels; recent matching-based objectives \citep{havens2025adjoint, liu2025adjoint} remove this memory cost but retain the on-policy training signal. Denoising regression has neither property: it is a simulation-free per-level regression whose data can be generated without limit from the unnormalized density.

\paragraph{Diffusion samplers with inner Monte Carlo.}
Another family of diffusion-inspired samplers replaces variational training with Monte Carlo estimation. iDEM \citep{akhoundsadegh2024idem} trains a network on Monte Carlo estimates of the noised score computed over a replay buffer, while RDMC \citep{huang2024reverse} estimates the denoising posterior mean by sampling and SLIPS \citep{grenioux2024stochastic} runs an inner MCMC chain at every noise level of an observation ladder, both at sampling time without a trained model. None of these applies an exact correction, which can lead to bias on complex targets. DiGS \citep{chen2024diffusive} is a relative which does have an exact correction: it noises the state to a single level and samples the denoising posterior with an inner Metropolis-corrected Langevin chain.

\paragraph{Exact MCMC with learned global proposals.}
DDMC belongs to a line of work that keeps MCMC exact while learning the global move. \citet{gabrie2022adaptive} use normalizing flows as MH proposals and demonstrate the repair of mode weights, and Ex$^2$MCMC \citep{samsonov2022local} composes global i-SIR proposals, optionally with adaptively trained flows, with local MALA refinement, the local--global pattern that the DDMC kernel \eqref{eq:ddmc} follows. Within this line, DDMC replaces the one-shot flow with a diffusion path. A flow proposal requires a tractable Jacobian determinant, tying exactness to invertible architectures, and must bridge noise and target in a single transformation; the path-space construction instead prices the proposal through Gaussian kernel densities, leaving the denoiser architecture unconstrained, and the ladder supplies the gradual annealing that a one-shot map must accomplish at once. Related transport approaches such as NeuTra \citep{hoffman2019neutra} instead reparameterize HMC with a flow rather than correcting a learned proposal.

%% file: sections/experiments.tex
\section{Experiments}
\label{sec:experiments}

In this section we apply DDMC to a variety of unnormalized densities. All samples in this section are initialized from the normal or uniform distribution that was used to create the MALA corpus. All chains are run for 10,000 steps across 4 independent seeds for each target. The first 400 steps of each chain are treated as burn-in samples and are not used when reporting metrics. The chains are thinned by taking every fourth sample after burn-in to obtain the final samples from the target density. Evaluations across all targets report the average acceptance ratio, IACT, and the energy Wasserstein-2 distance $E(\cdot)\,\mathcal{W}_2$, computed using the PDNS evaluation code \citep{guo2026proximal}; for particle targets we additionally report the sample Wasserstein-2 distance $\mathcal{W}_2$. See Appendix~\ref{app:metrics} for more information about metrics. $E(\cdot)\,\mathcal{W}_2$ and $\mathcal{W}_2$ are computed using reference samples from the literature when available and from our Parallel Tempering (PT) reference samples when unavailable. We also analyze mode occupancy and the probability of transitioning between modes when appropriate. We apply both the MH version of our algorithm from \eqref{eq:mhpath} and the i-SIR version of our algorithm from \eqref{eq:isirweight} in different scenarios. The results for the MH version are designated as DDMC and the results from the i-SIR version are designated as $\textrm{DDMC}_K$ where $K$ is the size of the i-SIR ensemble. Further details such as training and sampling configurations for each target can be found in Appendix~\ref{app:expdetails}.

\subsection{Mixture of Gaussians}

The first experiment is a proof of concept using the 2D reference distribution examined by iDEM \citep{akhoundsadegh2024idem} and DiGS \citep{chen2024diffusive}. This distribution has 40 evenly weighted mixture components which each have isotropic covariance. It is well-established that local samplers such as MALA and HMC cannot traverse between modes of this target in a realistic timeframe. A uniform distribution over the spread of the Gaussian modes is used to create the MALA corpus. Since MALA cannot cross between modes in this situation and since the corpus weighting is very dissimilar from the ground-truth weighting between modes due to the uniform initialization, this scenario is a proof-of-concept that DDMC can provide a correctly weighted proposal distribution via the MH correction even when its training distribution differs from the target.

The second experiment uses the bimodal Gaussian mixture proposed by \citet{grenioux2025improving}. We only examine their most challenging scenario with $d = 256$ dimensions and mode separation $a = 10$, where the two modes are centered at $\pm a \mathbf{1}_d$ and have diagonal covariances whose per-coordinate variances interpolate linearly between $0.01$ and $0.2$. The variances of each mixture component are opposed: the constrained direction of one mode corresponds to the least constrained direction of the other. Furthermore, the mixture has an uneven 1/3 and 2/3 weighting. Our training corpus from random initialization assigns about 50\% of mass in each mode, so this situation also allows us to examine whether DDMC can achieve the correct sample weighting between modes.

Results are shown in Table~\ref{tab:gmm}. In both cases we can achieve high acceptance rate, match the energy spectrum of the target density, and recover the weighting of the ground truth distribution within statistical uncertainty. This corroborates the claim that training on imbalanced MALA samples can still yield an effective global sampler. It is likely that more extreme weighting imbalances would have a negative effect on acceptance probability, and principled techniques for initializing the MALA corpus in a way that approximately covers separated modes are an important direction for future work.

\begin{table}[t]
\centering
\small
\caption{Gaussian mixture results. Both targets are evaluated against exact draws. The GT floor column is the exact-vs-exact $E(\cdot)\,\mathcal{W}_2$ value at $n = 2000$, i.e., the value attained when both sample sets are exact draws from the target: DDMC sits at the floor on both targets. The mode-weighting panel shows that the MH correction repairs the mode weights of the training corpus: the DDMC chain recovers the ground-truth weighting even though the corpus is far from it. Modes covered is the minimum over chains, so every chain covers every mode. For MoG-40 all ground-truth weights are equal, so the low-mode occupancy is the least-occupied mode: a minimum over 40 modes that is biased below $1/40$ even for exact draws, as the reference column shows. All DDMC entries lie within $2$ sd of exact-draw controls at matched effective sample size.}
\label{tab:gmm}
\begin{tabular}{l c c c c c}
\toprule
Target & $d$ & accept $\uparrow$ & IACT $\downarrow$ & $E(\cdot)\,\mathcal{W}_2 \downarrow$ & \emph{GT floor} \\
\midrule
MoG-40  & 2   & $0.724 \pm 0.004$ & $1.14 \pm 0.04$ & $0.11 \pm 0.03$ & \emph{$0.09 \pm 0.03$} \\
GMM-256 & 256 & $0.390 \pm 0.004$ & $3.89 \pm 0.73$ & $0.71 \pm 0.25$ & \emph{$0.65 \pm 0.13$} \\
\bottomrule
\end{tabular}

\vspace{0.6em}

\footnotesize
\setlength{\tabcolsep}{4pt}
\begin{tabular}{l c c c c c c}
\toprule
& & & & \multicolumn{3}{c}{Low-mode occupancy} \\
\cmidrule(lr){5-7}
Target & \makecell{Modes\\covered} & \makecell{TV\\(corpus)} & \makecell{TV\\(DDMC)} & DDMC & reference & \makecell{GT\\weight} \\
\midrule
MoG-40  & $40\,/\,40$ & $0.251$ & $0.074 \pm 0.017$ & $0.017 \pm 0.001$ & $0.018 \pm 0.001$ & $0.025$ \\
GMM-256 & $2\,/\,2$   & $0.165$ & $0.014 \pm 0.010$ & $0.345 \pm 0.014$ & $0.332 \pm 0.009$ & $0.333$ \\
\bottomrule
\end{tabular}
\end{table}

\subsection{Particle Potentials}

This section examines the performance of our sampler on targets which give the potential energy of particle ensembles. This setting allows us to examine targets with a more complex geometry than the Gaussian mixture experiments and allows direct comparison with a variety of diffusion-inspired methods from prior works. Following prior works, the architectures used in these experiments account for symmetries of the potentials, notably invariance to swapping of particle labels and equivariance to particle rotation. See Appendix~\ref{app:configs} for architecture details.

The results in Table~\ref{tab:particles-lit} show that DDMC can effectively model each of these target densities. The $E(\cdot)\,\mathcal{W}_2$ values of DDMC samples sit at or within uncertainty of the ground-truth floor of the published reference samples. The benefit of DDMC is particularly pronounced for LJ-55, which is the most complex target among the three. While DDMC does not match the DW-4 sample distance $\mathcal{W}_2$ of the published reference samples, we found that it does match an independent set of equilibrium reference samples generated by parallel tempering: against the PT reference the sample distance falls to $0.31 \pm 0.01$ for DDMC and $0.33 \pm 0.04$ for DDMC$_{33}$, at the measured ground-truth floor of $0.31 \pm 0.02$ for this statistic, while $E(\cdot)\,\mathcal{W}_2$ is unchanged within noise against either reference. This asymmetry indicates that the DW-4 sample-distance gap in Table~\ref{tab:particles-lit} reflects properties of the published reference rather than error of the sampler.

To demonstrate the benefit of using a diffusion denoiser rather than a variational objective for path proposals of the form \eqref{eq:mhpath}, we investigate the concurrent work MAD-Path \citep{chen2026madpath}. A MAD-Path score model is trained on LJ-55 using the project codebase. We find that the $\log \tilde{q}$ value obtained by MAD-Path samples across training settles around a negative value with a magnitude in the thousands and does not improve to reach the reference range around 325. The result is consistent with the poor performance of DDS and PIS on LJ-55 that was demonstrated in prior work, since these works use a closely related variational objective. This provides preliminary evidence that denoising diffusion leads to a more effective sampler than variational objectives for proposals using \eqref{eq:mhpath}. Both sample-based denoiser training and density-based variational training are potentially useful realizations of path-space MCMC, and understanding their relative strengths is a valuable direction for future research.

\begin{table}[t]
\centering
\small
\setlength{\tabcolsep}{4pt}
\caption{Evaluation on the particle-based energy functions, following the protocol and presentation of \citet{guo2026proximal}: Wasserstein-2 distances with respect to samples, $\mathcal{W}_2$ ($\downarrow$), and energies, $E(\cdot)\,\mathcal{W}_2$ ($\downarrow$), at $n = 2000$. Baseline values are as reported by \citet{guo2026proximal}; our values are mean $\pm$ sd across independent chains, computed with their released evaluation code against the same ground-truth samples. The final row gives the measured ground-truth-vs-ground-truth floor: values at the floor are statistically indistinguishable from exact samples at this sample size. Bold marks the best value in each column, including ties within reported uncertainty.}
\label{tab:particles-lit}
\footnotesize
\setlength{\tabcolsep}{2pt}
\begin{tabular}{l cc cc cc}
\toprule
& \multicolumn{2}{c}{DW-4 ($d{=}8$)} & \multicolumn{2}{c}{LJ-13 ($d{=}39$)} & \multicolumn{2}{c}{LJ-55 ($d{=}165$)} \\
\cmidrule(lr){2-3} \cmidrule(lr){4-5} \cmidrule(lr){6-7}
Method & $\mathcal{W}_2 \downarrow$ & $E(\cdot)\,\mathcal{W}_2 \downarrow$ & $\mathcal{W}_2 \downarrow$ & $E(\cdot)\,\mathcal{W}_2 \downarrow$ & $\mathcal{W}_2 \downarrow$ & $E(\cdot)\,\mathcal{W}_2 \downarrow$ \\
\midrule
PDDS \citep{phillips2024particle} & $0.92 \pm 0.08$ & $0.58 \pm 0.25$ & $4.66 \pm 0.87$ & $56.01 \pm 10.80$ & --- & --- \\
SCLD \citep{chen2025sequential} & $1.30 \pm 0.64$ & $0.40 \pm 0.19$ & $2.93 \pm 0.19$ & $27.98 \pm 1.26$ & --- & --- \\
PIS \citep{zhang2022path} & $0.68 \pm 0.28$ & $0.65 \pm 0.25$ & $1.93 \pm 0.07$ & $18.02 \pm 1.12$ & $4.79 \pm 0.45$ & $228.70 \pm 131.27$ \\
DDS \citep{vargas2023denoising} & $0.92 \pm 0.11$ & $0.90 \pm 0.37$ & $1.99 \pm 0.13$ & $24.61 \pm 8.99$ & $4.60 \pm 0.09$ & $173.09 \pm 18.01$ \\
LV-PIS \citep{richter2024improved} & $1.04 \pm 0.29$ & $1.89 \pm 0.89$ & --- & --- & --- & --- \\
iDEM \citep{akhoundsadegh2024idem} & $0.70 \pm 0.06$ & $0.55 \pm 0.14$ & $1.61 \pm 0.01$ & $30.78 \pm 24.46$ & $4.69 \pm 1.52$ & $93.53 \pm 16.31$ \\
AS \citep{havens2025adjoint} & $0.62 \pm 0.06$ & $0.55 \pm 0.12$ & $1.67 \pm 0.01$ & $2.40 \pm 1.25$ & $4.50 \pm 0.05$ & $58.04 \pm 20.98$ \\
ASBS \citep{liu2025adjoint} & $\mathbf{0.38 \pm 0.05}$ & $0.19 \pm 0.03$ & $1.59 \pm 0.00$ & $1.28 \pm 0.22$ & $4.00 \pm 0.03$ & $27.69 \pm 3.86$ \\
PDNS \citep{guo2026proximal} & $0.51 \pm 0.04$ & $0.21 \pm 0.03$ & $\mathbf{1.57 \pm 0.01}$ & $1.01 \pm 0.18$ & $3.95 \pm 0.01$ & $21.97 \pm 3.14$ \\
\midrule
DDMC (ours) & $0.50 \pm 0.05$ & $\mathbf{0.14 \pm 0.04}$ & $1.59 \pm 0.01$ & $\mathbf{0.42 \pm 0.09}$ & $3.91 \pm 0.00$ & $\mathbf{1.22 \pm 0.43}$ \\
DDMC$_K$ (ours) & $0.50 \pm 0.05$ & $\mathbf{0.14 \pm 0.04}$ & $\mathbf{1.58 \pm 0.00}$ & $\mathbf{0.42 \pm 0.10}$ & $\mathbf{3.88 \pm 0.00}$ & $\mathbf{0.83 \pm 0.31}$ \\
\midrule
GT floor & $0.31 \pm 0.02$ & $0.12 \pm 0.03$ & $1.44 \pm 0.01$ & $0.40 \pm 0.13$ & $2.57 \pm 0.04$ & $0.87 \pm 0.23$ \\
\bottomrule
\end{tabular}
\end{table}

\begin{table}[t]
\centering
\small
\caption{DDMC chain statistics on particle potentials.}
\label{tab:particles-chain}
\begin{tabular}{l l c c}
\toprule
Target & Sampler & accept $\uparrow$ & IACT $\downarrow$ \\
\midrule
DW-4  & DDMC          & $0.545 \pm 0.005$ & $1.30 \pm 0.08$ \\
      & DDMC$_{33}$   & $0.941 \pm 0.002$ & $1.03 \pm 0.04$ \\
\midrule
LJ-13 & DDMC          & $0.277 \pm 0.006$ & $5.37 \pm 0.94$ \\
      & DDMC$_{33}$   & $0.831 \pm 0.005$ & $1.37 \pm 0.04$ \\
\midrule
LJ-55 & DDMC          & $0.028 \pm 0.003$ & $29.8 \pm 6.1$ \\
      & DDMC$_{64}$   & $0.342 \pm 0.004$ & $4.67 \pm 0.46$ \\
\bottomrule
\end{tabular}
\end{table}

\subsection{Bayesian Neural Network}

The last experiment covers the 550-dimensional Bayesian Neural Network (BNN) target examined by DiGS~\citep{chen2024diffusive}. We find that the posterior is effectively covered by local MALA sampling. The BNN target therefore serves as a stress test of trainability and exactness at scale rather than of mixing, for which the failure of local samplers is well-documented even in simple situations. The architecture used for this experiment was a DiT~\citep{peebles2023scalable} that respects the permutation invariance of the hidden labels. This target was the most difficult learning scenario that we examined and required significantly more MALA training samples than the other targets. We saw evidence that a fixed MALA corpus of 6.4M samples eventually led to a model which could no longer improve $\log r$. In particular, $\dpath$ drifted in a negative direction and erased gains from $\dq$. To avoid data exhaustion, we trained the DDMC BNN denoiser concurrently with producer workers who were continuously generating new MALA samples. This experiment led to a model which could obtain reasonable acceptance probabilities for global proposals whose samples match the spectrum of a PT reference, as shown in Table~\ref{tab:bnn}.

\begin{table}[t]
\centering
\small
\caption{DDMC results on the BNN posterior ($d = 550$, $K = 32$), evaluated against the independent pooled parallel tempering reference. The $T = 640$ row reports mean $\pm$ sd across six independent runs; $T = 2560$ is a single long run. The GT floor column is the PT-vs-PT $E(\cdot)\,\mathcal{W}_2$ value at $n = 2000$: the six-run mean sits below the floor, and five of the six runs are individually at or below it.}
\label{tab:bnn}
\begin{tabular}{l c c c c}
\toprule
Sampler & accept $\uparrow$ & IACT $\downarrow$ & $E(\cdot)\,\mathcal{W}_2 \downarrow$ & \emph{GT floor} \\
\midrule
DDMC$_{32}$ ($T{=}640$)  & $0.132 \pm 0.004$ & $4.87 \pm 0.27$ & $1.03 \pm 0.26$ & \emph{$1.17 \pm 0.27$} \\
DDMC$_{32}$ ($T{=}2560$) & $0.275$ & $3.61$ & $1.37$ & \emph{$1.17 \pm 0.27$} \\
\bottomrule
\end{tabular}
\end{table}

This BNN target was originally studied by DiGS, and we compare against DiGS and tuned local samplers on burn-in: the number of steps required to reach the equilibrium energy band from the cold initialization (Table~\ref{tab:burnin}). The comparison isolates the difference between global and local moves from a cold start. The first DDMC proposal is accepted with probability near one and carries the chain directly into the equilibrium band, so convergence takes one to two proposal rounds, $2.8$--$5.5\times$ faster than tuned MALA and $9$--$18\times$ faster than DiGS for a single chain. Running the exact released DiGS implementation at its own hyperparameters, we observe negligible acceptance of its global jump throughout sampling ($< 10^{-4}$), so its auxiliary-variable move contributes nothing at this dimension and its burn-in proceeds through its Langevin component alone. The DiGS and MALA rows of Table~\ref{tab:burnin} therefore compare the same underlying local sampler under different settings: our MALA baseline uses a step size adapted to this target, while DiGS runs with its original parameters and pays the per-step cost of proposing jumps that are never accepted, which explains the gap between the rows. The efficiency comparison depends on the workload: DDMC is the more efficient choice when a small number of chains is needed or when chains are spread across many parallel workers, while a single device batching hundreds of chains favors MALA, whose inexpensive energy evaluations sit far below GPU saturation at batch one and amortize across chains nearly for free (final column of Table~\ref{tab:burnin}). We emphasize that the central accomplishment on this target is not throughput but the existence of the proposal itself: exact global proposals accepted at rates above $10\%$ in $550$ dimensions, where global moves without an accurate model are essentially never accepted. We also repeated the test-NLL evaluation from the original DiGS protocol; under that protocol no pair of methods can be statistically separated, and we report it in Appendix~\ref{app:nll} for completeness.

\begin{table}[t]
\centering
\small
\caption{Burn-in to the equilibrium energy band on the BNN posterior ($d = 550$). Chains per method start from the cold initialization and a chain is said to be converged when its energy distribution enters the pooled PT reference band. Wall-clock convergence time for 1 chain and a batch of 512 parallel chains on a single A6000 GPU are shown.}
\label{tab:burnin}
\begin{tabular}{l c c c}
\toprule
Method & to converge & 1 chain & 512 chains \\
\midrule
DDMC ($T{=}640$) & $1$--$2$ rounds & $18$--$36$ s & $403$--$806$ s \\
MALA (adapted step) & $71{,}200$ steps & $99$ s & $130$ s \\
DiGS & $129{,}800$ steps & $330$ s & $356$ s \\
\bottomrule
\end{tabular}
\end{table}

%% file: sections/conclusion.tex
\section{Conclusion}
\label{sec:conclusion}

This paper demonstrates that a model trained using denoising diffusion can be directly utilized as an exact global MCMC sampler. Training on local MALA samples and composing the global path proposal with a local MALA proposal leads to a natural synergy between the proposal components where global path proposals obtain states similar to the training distribution from the MALA updates while overcoming the limited local mixing that is typical of MALA and local samplers. We call this composed sampler DDMC. MH correction ensures that chains remain exact even when the MALA corpus mode weighting differs from the ground truth. Experiments across a variety of target densities demonstrate that DDMC can achieve global proposals with high acceptance while adhering closely to the target distribution. The DDMC framework also provides a straightforward mechanism for MCMC samplers to benefit from the scalability of denoising diffusion, which has been extensively studied in the image domain and related applications. Limitations of DDMC include the computational cost of reverse diffusion paths and susceptibility to modeling failure in cases where MALA samples do not cover, or greatly underrepresent, high-probability regions. Reducing sampling computational cost and providing mechanisms to create MALA corpora whose mode weightings are reasonable approximations of the target are important directions for future work.

%% file: sections/appendix.tex
\section{Ergodicity of the Noise-to-Denoise Transition}
\label{app:ergodicity}

The general principle: a Markov kernel that leaves $q$ invariant and whose transition distribution has a component with a density positive at every pair $(x, x')$ with $\tilde{q}(x') > 0$ is $q$-irreducible and aperiodic, and therefore ergodic for $q$ \citep{tierney1994markov, mengersen1996rates}. It remains to check positivity in each case.

\paragraph{SDE case.}
The forward step $Y = X + \sigma Z$ has the everywhere-positive density $\N(y; x, \sigma^2 I_d)$, and the reverse pass composes Gaussian kernels with strictly positive variances, so the composite transition density is strictly positive on $\R^d \times \R^d$. The learned means can move the kernels anywhere but cannot remove support.

\paragraph{ODE case.}
For every $\sigma > 0$, $q_\sigma = q * \N(0, \sigma^2 I_d)$ is smooth with a strictly positive density, so the probability-flow velocity field is smooth; by Tweedie's identity \citep{efron2011tweedie} it equals $(x - \E[X \mid x])/\sigma$ up to the schedule factor, which has at most linear growth whenever $q$ has sub-Gaussian tails, ruling out finite-time blow-up. The reverse flow between any two positive noise levels is then a diffeomorphism of $\R^d$, and the pushforward of the full-support forward Gaussian through a diffeomorphism retains an everywhere-positive density. Unlike image modeling, where data concentrate near a lower-dimensional manifold, the targets considered here have effective dimension $d$. Thermal fluctuations give the density full-dimensional support, so no degeneracy arises at small $\sigma$.

\paragraph{DDMC transition.}
Each component of the DDMC cycle leaves $q$ invariant, so the cycle does. The forward increments \eqref{eq:fwdkernel} and reverse kernels \eqref{eq:revkernel} are positive-variance Gaussians and the acceptance probability \eqref{eq:mhpath} is strictly positive wherever $\tilde{q} > 0$, so the accepted-move component of the cycle satisfies the positivity condition of the general principle above, and the DDMC chain is ergodic for $q$.

\section{Experimental Details}
\label{app:expdetails}

\subsection{Per-Target Configurations}
\label{app:configs}

Table~\ref{tab:configs} summarizes the configuration used for each target. Full training configurations, corpus-generation schedules, and evaluation scripts will be released with the code.

\begin{table}[t]
\centering
\small
\setlength{\tabcolsep}{4pt}
\caption{Per-target configuration. ``Corpus'' is the number of MALA training samples; for the BNN the corpus is refreshed continuously during training from a 6.4M-state seed. ``Steps'' is the training step of the shipped checkpoint. $K$ is the i-SIR pool size where pooling is used, $N$ is the number of MALA steps per DDMC cycle, and $n_\textrm{cal}$ is the number of held-out states used to calibrate the reverse variances.}
\label{tab:configs}
\begin{tabular}{l r r l r r r r r r r r}
\toprule
Target & $d$ & Corpus & Architecture & Params & Steps & $T$ & $\sigma_{\max}$ & $\sigma_{\min}$ & $K$ & $N$ & $n_\textrm{cal}$ \\
\midrule
MoG-40  & 2   & 10M  & MLP ($512 \times 6$)               & 3M & 60K  & 320 & 0.38  & 0.005   & ---     & 20 & 3072 \\
GMM-256 & 256 & 2M   & token MLP ($128 \times 4$)         & 0.9M     & 50K  & 320 & 18.92 & 0.003   & ---     & 20 & 3072 \\
DW-4    & 8   & 4M   & equivariant DiT ($256 \times 6$)   & 7.5M     & 400K & 40  & 1.62  & 0.05  & 33      & 20 & 3072 \\
LJ-13   & 39  & 4M   & equivariant DiT ($256 \times 6$)   & 7.5M     & 360K & 80  & 2.93  & 0.02  & 33      & 20 & 3072 \\
LJ-55   & 165 & 6M   & equivariant DiT ($256 \times 6$)   & 7.5M     & 580K & 320 & 4.0   & 0.005 & 64      & 20 & 2048 \\
BNN     & 550 & 6.4M & DiT ($1024 \times 16$)        & 305M     & 1.4M & 640/2560 & 20.0 & 0.003 & 32 & 1000 & 3072 \\
\bottomrule
\end{tabular}
\end{table}

\subsection{Metrics}
\label{app:metrics}

The energy Wasserstein-2 distance $E(\cdot)\,\mathcal{W}_2$ and the sample Wasserstein-2 distance $\mathcal{W}_2$ are computed with the evaluation code released by \citet{guo2026proximal}, unmodified, at $n = 2000$ samples per chain. All metrics are computed per chain and never pooled across chains; the reported uncertainty is the standard deviation across independent chains. Ground-truth floors are measured by evaluating the same metric between two independent reference sample sets at the same $n$: disjoint splits of a stored reference bank, or fresh exact draws when the target is analytically samplable, averaged over 20 to 50 replicate pairs. The floor is the value of the metric when both sample sets are exact, so results at the floor are statistically indistinguishable from exact sampling at this sample size, and rankings between values inside the floor are not meaningful. The integrated autocorrelation time (IACT) is estimated on the raw energy trace of each chain, in rounds, before thinning, using the initial-positive-sequence estimator of \citet{geyer1992practical}. IACT measures the number of chain steps needed to obtain one effectively independent sample, so a trace of length $n$ carries roughly $n / \textrm{IACT}$ independent samples and a value near one indicates nearly independent consecutive states.

\subsection{Reference Samples}
\label{app:references}

For MoG-40 and GMM-256 the reference is exact draws from the analytically samplable target. For the particle targets the published reference is the iDEM/PDNS ground-truth test split for each system; we verified that the DW-4 split shipped with the PDNS code is byte-identical to iDEM's. For the BNN the energy marginal is evaluated against a held-out bank of $4{,}096$ MALA states that is disjoint from both the training corpus and the calibration bank.

We additionally constructed independent parallel tempering references. DW-4: 12 temperatures ($\beta$ from $1.0$ to $0.05$), $1{,}536{,}000$ states after burn-in, replica swap acceptance $0.72$--$0.88$. LJ-55: 24 rungs ($\beta$ from $1$ to $0.10$), $96{,}000$ states, $217$ hot-to-cold traversals, swap acceptance $0.23$--$0.57$. BNN: seven independent PT runs pooled, each with a 40-rung likelihood-only ladder, totaling $2{,}748{,}992$ states with $9{,}632$ hot-to-cold traversals and minimum swap acceptance $0.33$.

\subsection{Test NLL Under the DiGS Protocol}
\label{app:nll}

The original DiGS evaluation on the BNN target reports test negative log-likelihood under a fixed compute protocol: two chains, 75 recorded samples with 5{,}000 gradient steps between recordings. We repeat this protocol for five methods across four seeds, running DiGS with its exact released implementation; for DDMC, a single path proposal moves the chain to the equilibrium region (its wall-clock cost is converted to MALA steps and charged against the budget) and MALA is run for the remainder. Results are in Table~\ref{tab:nll}. No pairwise Welch test separates any two methods once multiplicity is accounted for: eight of ten comparisons have $p \geq 0.11$, and the lowest ($p = 0.029$, DDMC vs.\ DiGS) does not survive a Bonferroni correction at $p < 0.005$. We therefore make no claim that any method outperforms another under this metric. Test NLL is not a sampling-fidelity metric, since it can reward under-dispersed samples, which is why this comparison is reported here rather than in the main text.

\begin{table}[t]
\centering
\small
\caption{Test NLL on the BNN target under the protocol of \citet{chen2024diffusive} (mean $\pm$ sd across 4 seeds, expected-NLL estimator). No pair of methods is statistically separable after correcting for multiple comparisons.}
\label{tab:nll}
\begin{tabular}{l c c}
\toprule
Method & test NLL $\downarrow$ & wall-clock \\
\midrule
DiGS & $0.229 \pm 0.020$ & $1032$ s \\
MALA (DiGS step size) & $0.237 \pm 0.038$ & $852$ s \\
MALA (tuned step size) & $0.208 \pm 0.037$ & $740$ s \\
HMC & $0.216 \pm 0.025$ & $526$ s \\
DDMC (one proposal $+$ MALA) & $0.194 \pm 0.011$ & $787$ s \\
\bottomrule
\end{tabular}
\end{table}

\subsection{Compute}
\label{app:compute}

All experiments ran on NVIDIA A6000 GPUs with PyTorch. Path log-densities are accumulated in double precision to obtain a numerically accurate MH acceptance probability. The largest model (BNN, 305M parameters) was trained on two A6000 GPUs for approximately $280$ GPU-hours, with corpus production running concurrently on the remaining GPUs of the node. The particle and mixture models ($0.9$M--$7.5$M parameters) are one to two orders of magnitude cheaper to train. Sampling for all reported chains used one GPU per chain.

\section{Diagnostic Terms During Training}
\label{app:diagnostics}

The figures below show the diagnostic quantities of Section~\ref{sec:diagnostics} over training for each target: histograms of $\dq$, $\dpath$, and $\log r$, together with the evolution of their means and of the acceptance $\bar{\alpha}$ \eqref{eq:diagacc}. The sampling configurations used for these diagnostics were chosen for monitoring during training and may differ from the inference settings of Section~\ref{sec:experiments}; the trends are similar across configurations even when the $\log r$ shown in a diagram is not the optimal one. Two empirical observations stand out. First, $\dpath$ tends to remain clustered around zero throughout healthy training: the forward and reverse path densities balance early, and the improvement in $\log r$ comes from $\dq$, that is, from the diffusion placing its proposal mass closer and closer to the true target. This behavior deserves a closer investigation, which we leave to future work. Second, $\dq$ and $\dpath$ have a strong negative correlation, which causes the spread of $\log r$ to be much tighter than the spread of either component.

\begin{figure}[ht]
\centering
\includegraphics[width=\textwidth]{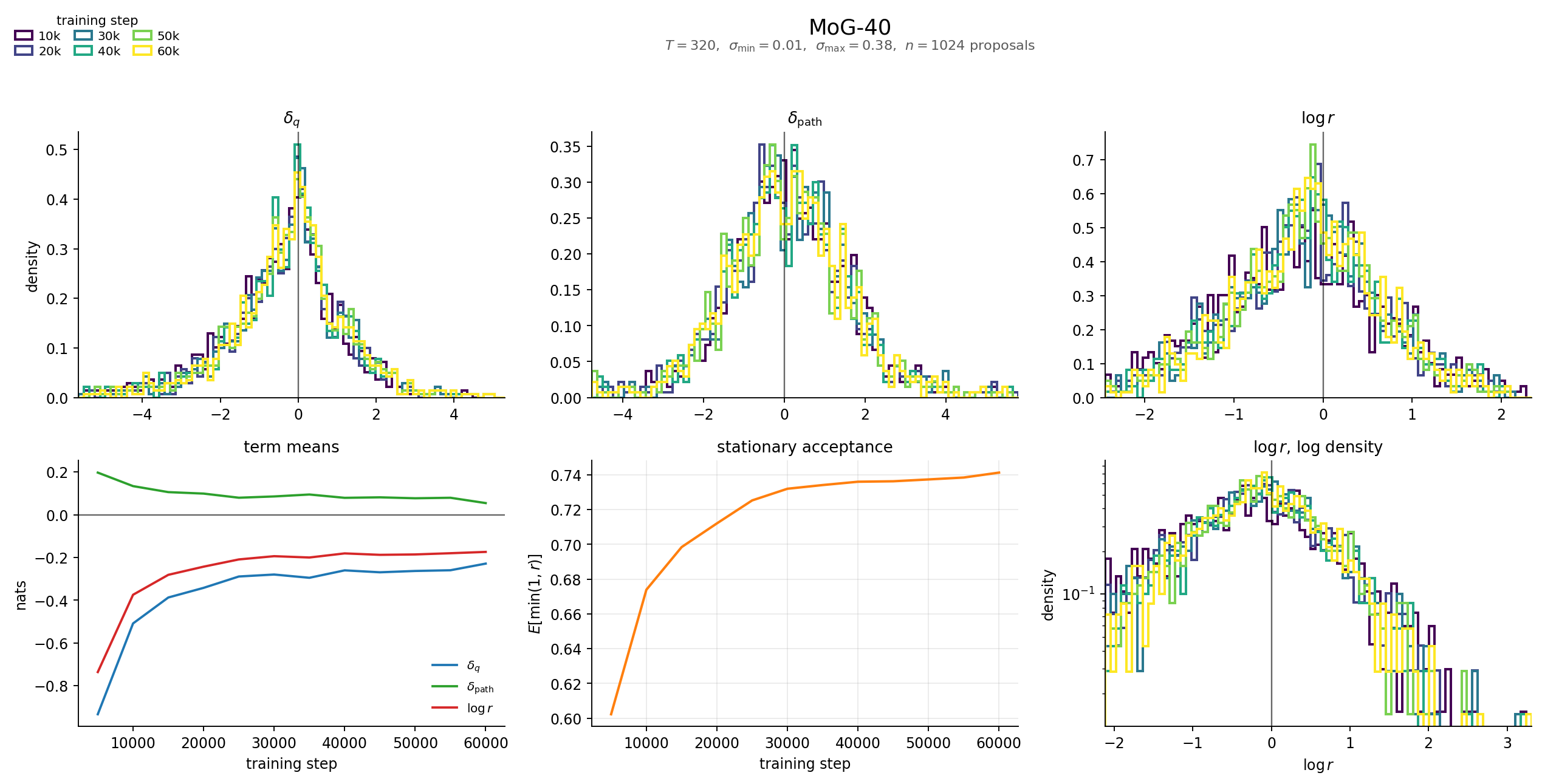}
\caption{Training diagnostics for MoG-40.}
\label{fig:score-mog}
\end{figure}

\begin{figure}[ht]
\centering
\includegraphics[width=\textwidth]{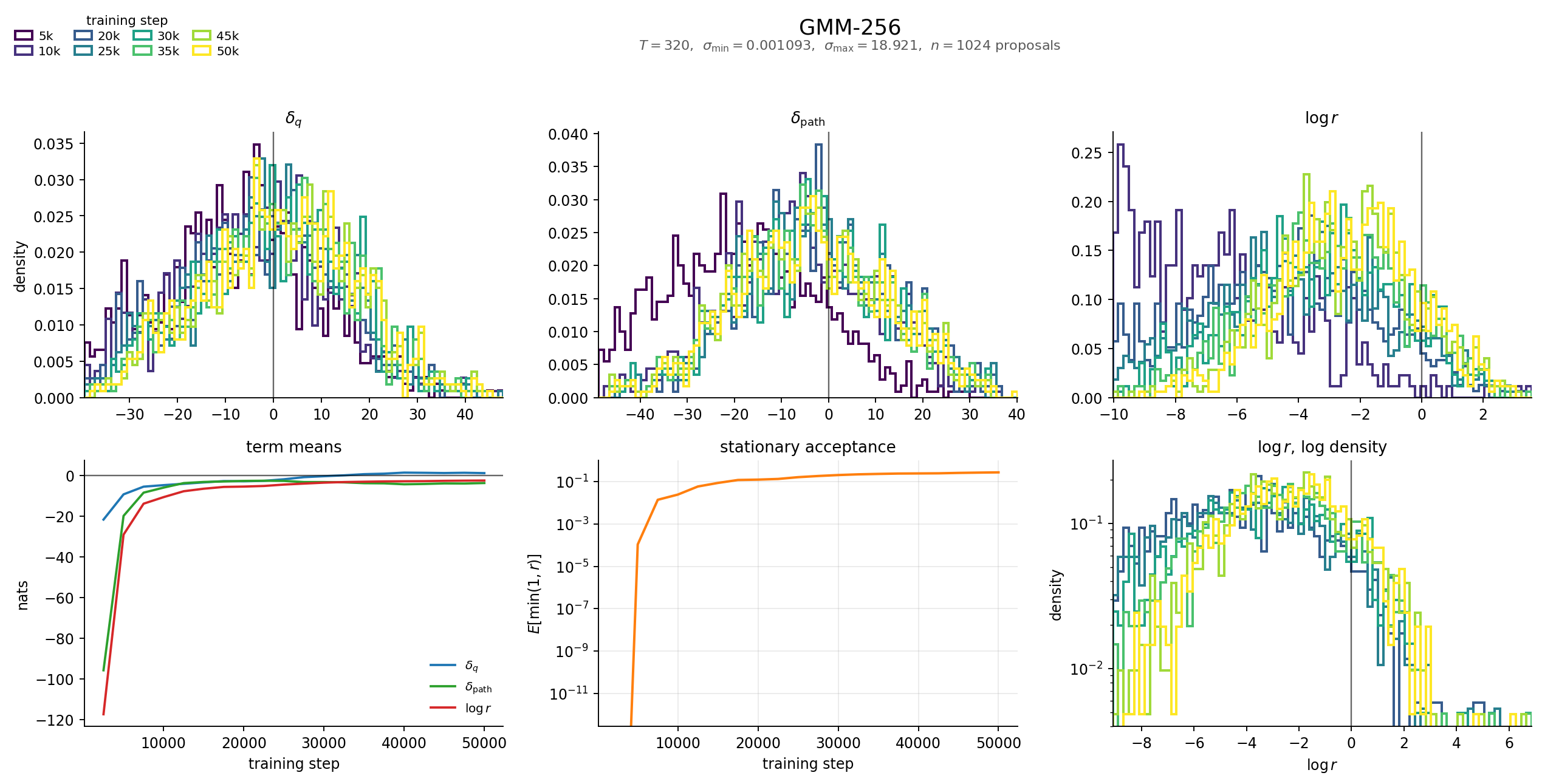}
\caption{Training diagnostics for GMM-256.}
\label{fig:score-gmm}
\end{figure}

\begin{figure}[ht]
\centering
\includegraphics[width=\textwidth]{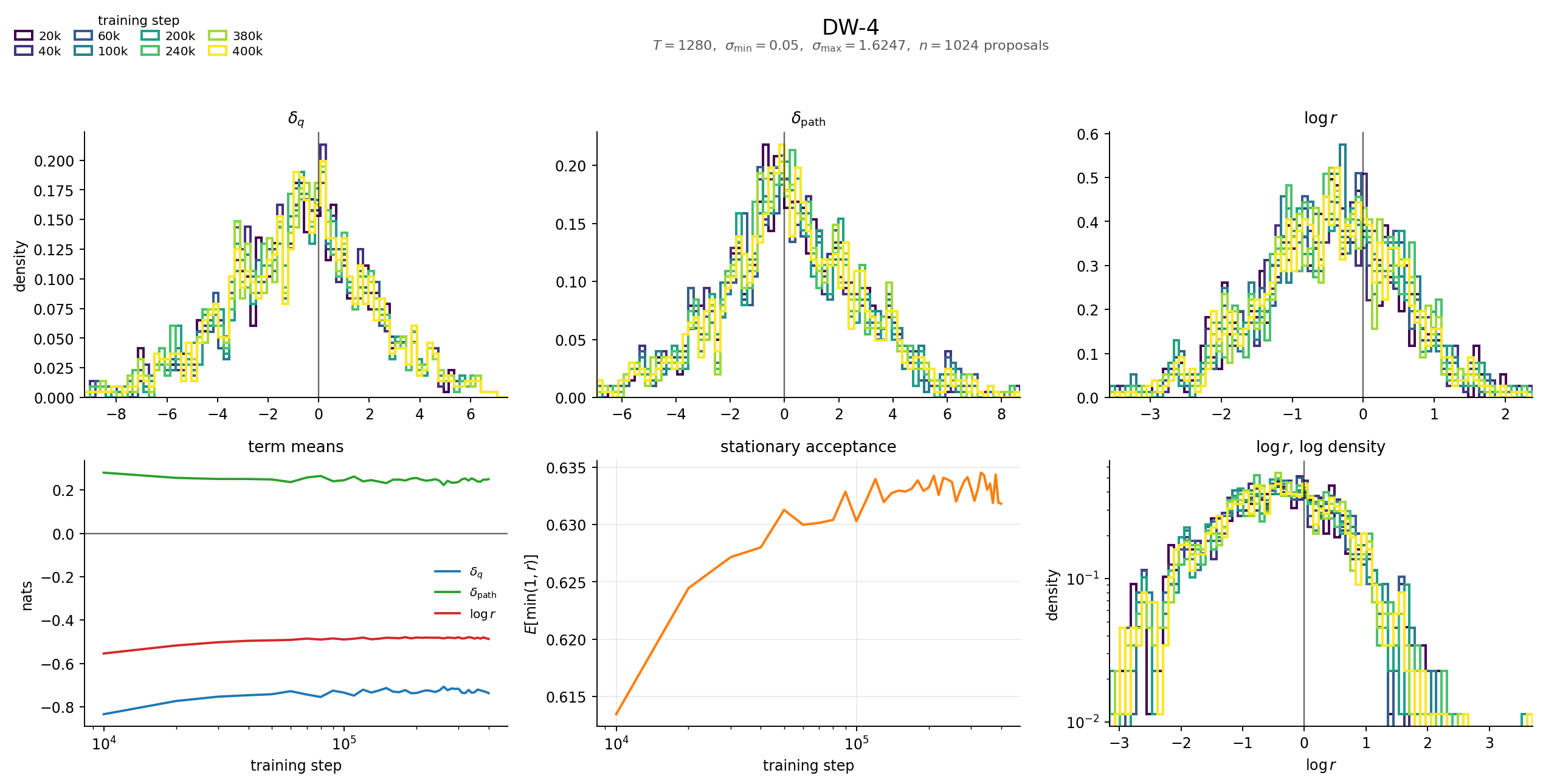}
\caption{Training diagnostics for DW-4.}
\label{fig:score-dw4}
\end{figure}

\begin{figure}[ht]
\centering
\includegraphics[width=\textwidth]{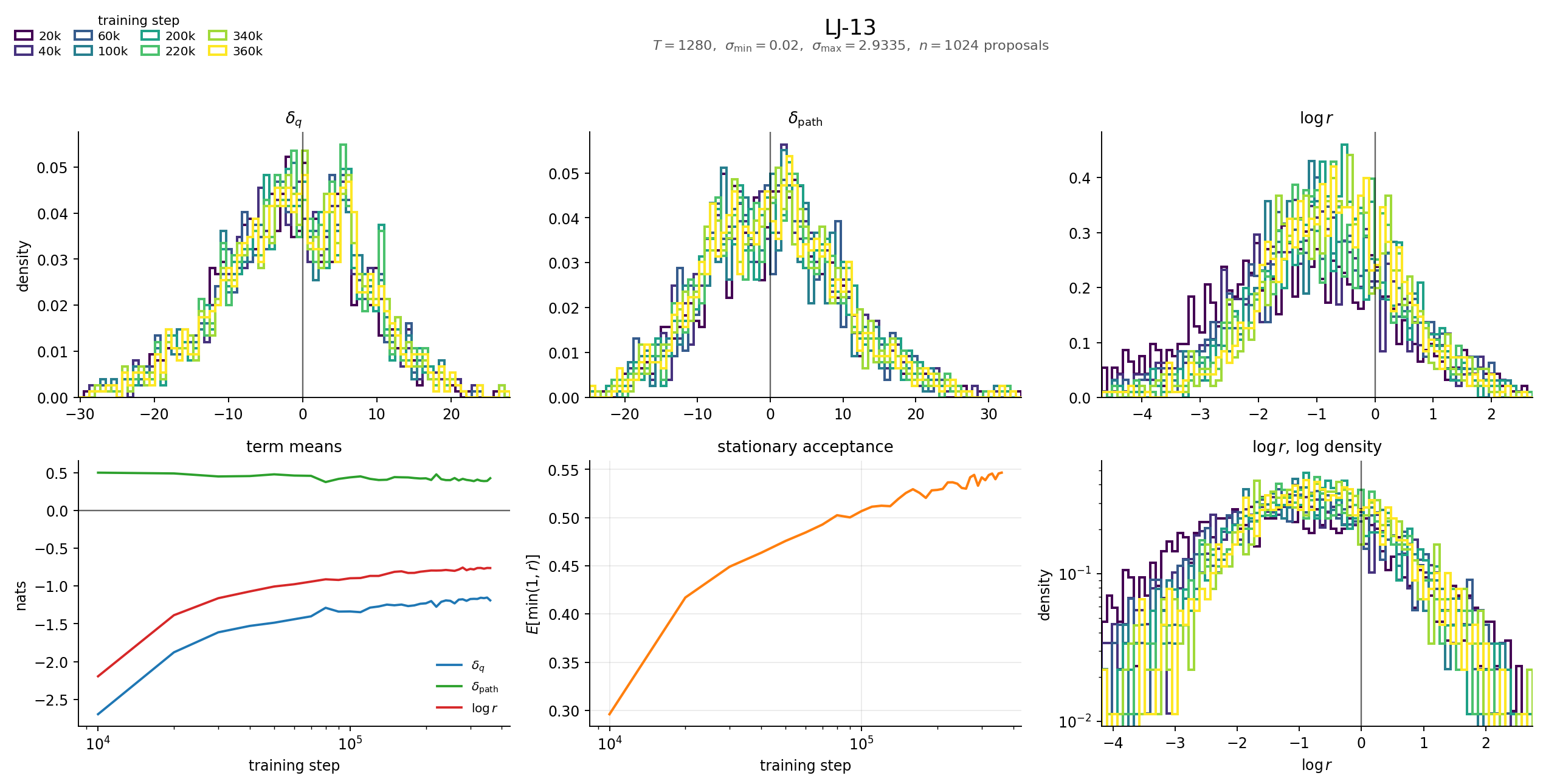}
\caption{Training diagnostics for LJ-13.}
\label{fig:score-lj13}
\end{figure}

\begin{figure}[ht]
\centering
\includegraphics[width=\textwidth]{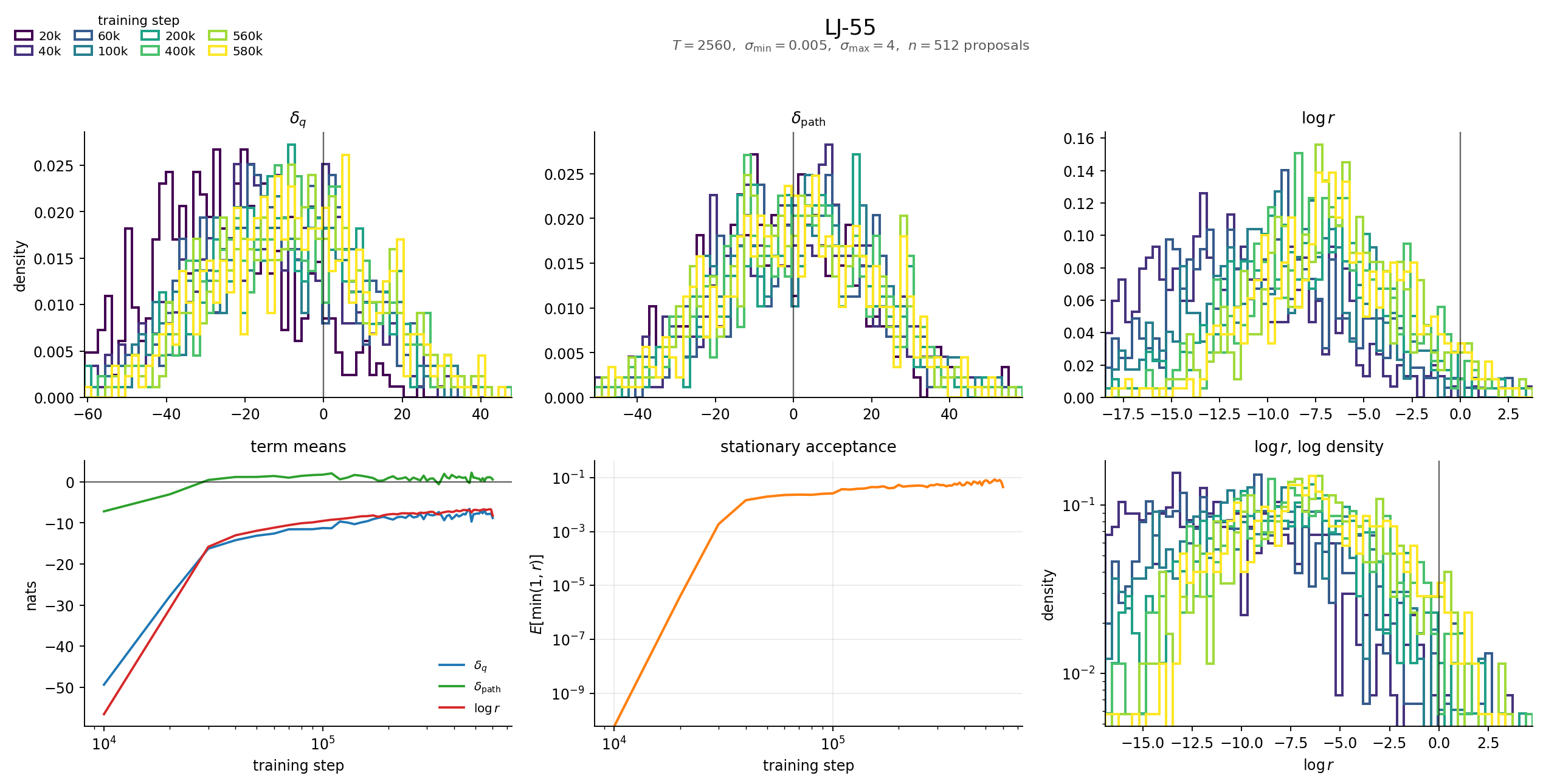}
\caption{Training diagnostics for LJ-55.}
\label{fig:score-lj55}
\end{figure}

\begin{figure}[ht]
\centering
\includegraphics[width=\textwidth]{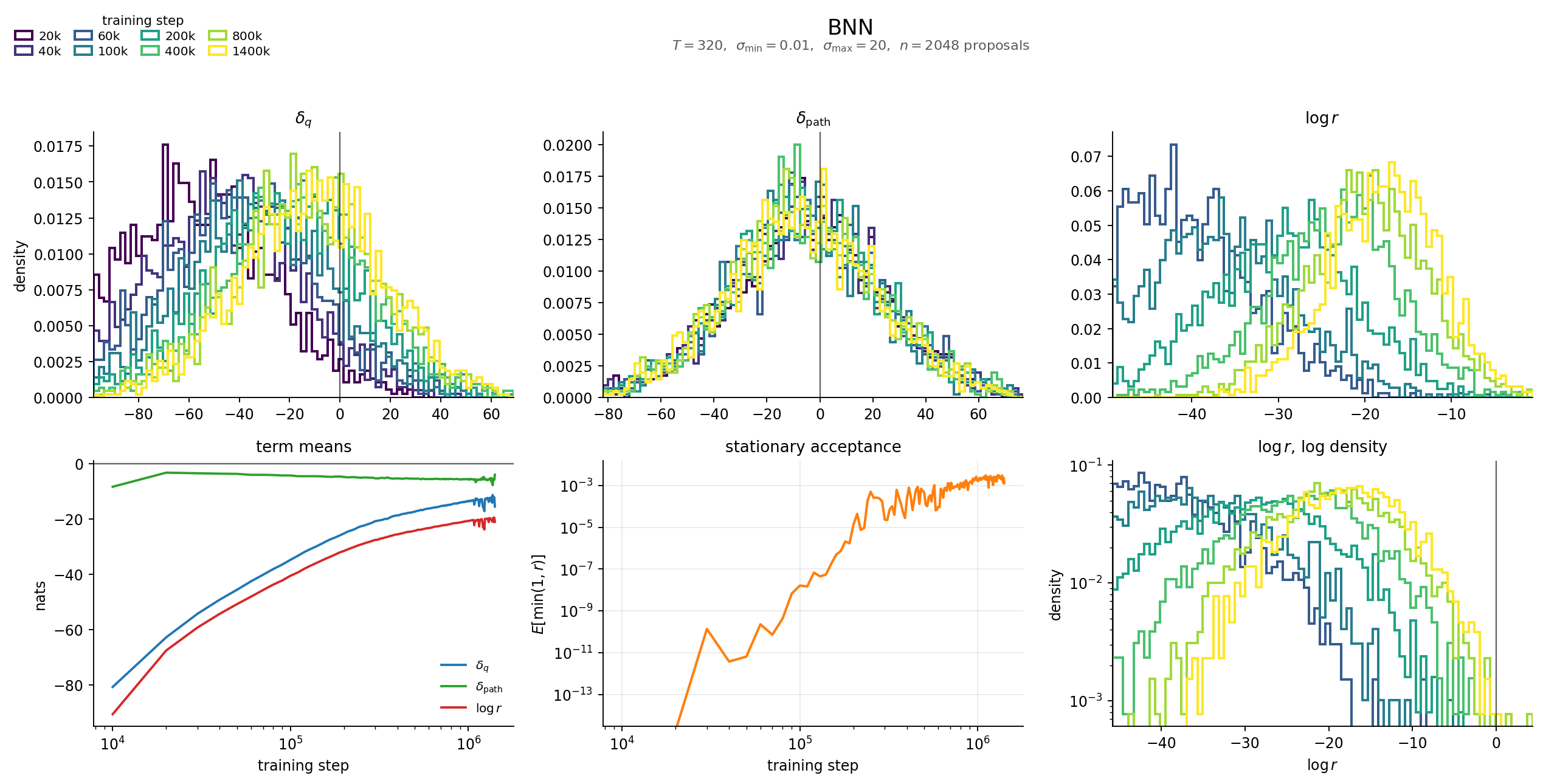}
\caption{Training diagnostics for the BNN posterior.}
\label{fig:score-bnn}
\end{figure}